\documentclass[10pt,twocolumn,letterpaper]{article}
\usepackage[pagenumbers]{iccv} %

\usepackage{multirow}
\definecolor{iccvblue}{rgb}{0.21,0.49,0.74}
\usepackage[pagebackref,breaklinks,colorlinks,allcolors=iccvblue]{hyperref}
\def\paperID{*****} %
\def\confName{ICCV}
\def\confYear{2025}
\title{Towards Suturing World Models:\\Learning Predictive Models for Robotic Surgical Tasks}
\author{
Mehmet Kerem Turkcan$^{1}$ \quad Mattia Ballo$^{2}$ \quad Filippo Filicori$^{2}$ \quad Zoran Kostic$^{1}$\\
\\
$^{1}$Columbia University, New York, NY, USA\\
$^{2}$Northwell Health, Lenox Hill Hospital, New Hyde Park, NY, USA\\
\\
\texttt{\{mkt2126,zk2172\}@columbia.edu, \{mballo,ffilicori\}@northwell.edu}
}

\begin{document}
\maketitle
\begin{abstract}
We introduce specialized diffusion-based generative models that capture the spatiotemporal dynamics of fine-grained robotic surgical sub-stitch actions through supervised learning on annotated laparoscopic surgery footage. The proposed models form a foundation for data-driven world models capable of simulating the biomechanical interactions and procedural dynamics of surgical suturing with high temporal fidelity. Annotating a dataset of $\sim2K$ clips extracted from simulation videos, we categorize surgical actions into fine-grained sub-stitch classes including ideal and non-ideal executions of needle positioning, targeting, driving, and withdrawal. We fine-tune two state-of-the-art video diffusion models, LTX-Video and HunyuanVideo, to generate high-fidelity surgical action sequences at $\ge$768×512 resolution and $\ge$49 frames. For training our models, we explore both Low-Rank Adaptation (LoRA) and full-model fine-tuning approaches. Our experimental results demonstrate that these world models can effectively capture the dynamics of suturing, potentially enabling improved training simulators, surgical skill assessment tools, and autonomous surgical systems. The models also display the capability to differentiate between ideal and non-ideal technique execution, providing a foundation for building surgical training and evaluation systems. We release our models for testing and as a foundation for future research. Project Page: \url{https://mkturkcan.github.io/suturingmodels/}
\end{abstract}
    
\begin{figure*}[h!]
  \centering
  \includegraphics[width=0.95\linewidth]{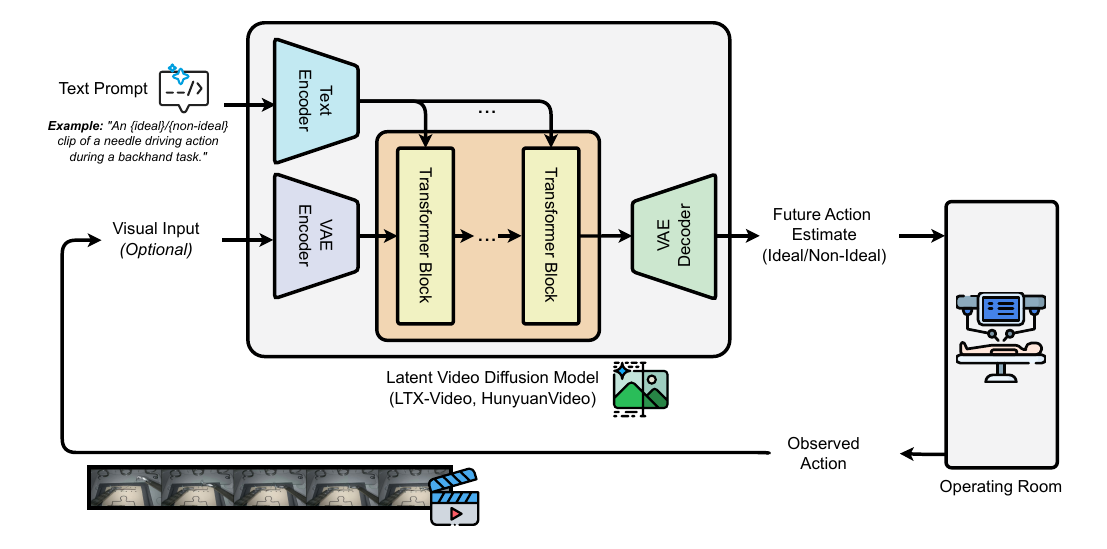} 
  \caption{Workflow of the proposed approach. We prompt latent video diffusion models with text and (optionally) visual input. The models are trained with expert-annotated ideal and non-ideal demonstrations, allowing for the models to output either class of quality.}
  \label{fig:surgery-workflow}
\end{figure*}

\section{Introduction}
\label{sec:intro}

Robotic-assisted minimally invasive surgery has revolutionized surgical practice by providing enhanced precision, control, and visualization during complex procedures, effectively reducing the learning curve despite the lack of haptic feedback \cite{sheetz2020trends,shaw2018robotic}.

However, robotic surgical training faces several challenges, including dependence on observational learning, low-quality feedback, and inconsistent assessment methods \cite{brian2024artificial}. To address these issues, the development of AI models for automated assessment has been proposed as a potential solution. Yet, this approach is hindered by the need for high-quality annotated data, which can be difficult to obtain. Events requiring documentation are often rare, and the process of labeling by expert surgeons is time-consuming, making it not always feasible, especially at scale.

World models are predictive frameworks that estimate environmental state transitions based on learned representations, and have shown effectiveness in a wide range of domains ranging from reinforcement learning in simulated environments to robotic control systems. Although world models could revolutionize surgical training, planning, and intraoperative guidance, their applications in these areas remain largely unexplored. In this work, we present a step towards ``Suturing World Models'', predictive models of surgical suturing tasks from video data that replicate physically accurate suturing actions. We give an overview of our proposed methodology in Figure \ref{fig:surgery-workflow}.

We use a granular annotation scheme that enables our models to learn the physics of suturing actions and distinguish variations in technique quality, vital for educational applications and autonomous surgical systems.

We leverage architectural innovations from state-of-the-art open-source video diffusion models, specifically LTX-Video (2B parameters) and HunyuanVideo (13B parameters), adapting them to the surgical domain through fine-tuning \cite{hacohen2024ltx,kong2024hunyuanvideo}. We fine-tune these models on a diverse dataset of laparoscopic surgical footage. 

We employ two distinct fine-tuning strategies in our work, full-parameter fine-tuning and low-rank adaptation (LoRA), enabling adaptation of large-scale pretrained models to the surgical domain and simultaneously minimizing computational requirements \cite{hu2022lora}. All models support high spatiotemporal resolution at $\ge$768×512 resolution with at least 49 frames per video, capturing the temporal dynamics of surgical actions with sufficient spatial detail for modeling complete sub-stitch actions.

Our contributions are threefold: (i) we introduce a novel application of video diffusion models to surgical technique modeling, demonstrating efficacy in capturing complex robotic suturing dynamics; (ii) we evaluate different fine-tuning strategies for video diffusion models in the surgical domain, (iii) the models we build can generate synthetic data for downstream applications for which collection of annotated real data could be extremely time-consuming.

Our work lays the foundation for next-generation surgical simulation, skill assessment, and potentially autonomous surgical systems.

\begin{figure*}[t!]
  \centering
  \includegraphics[width=0.99\linewidth]{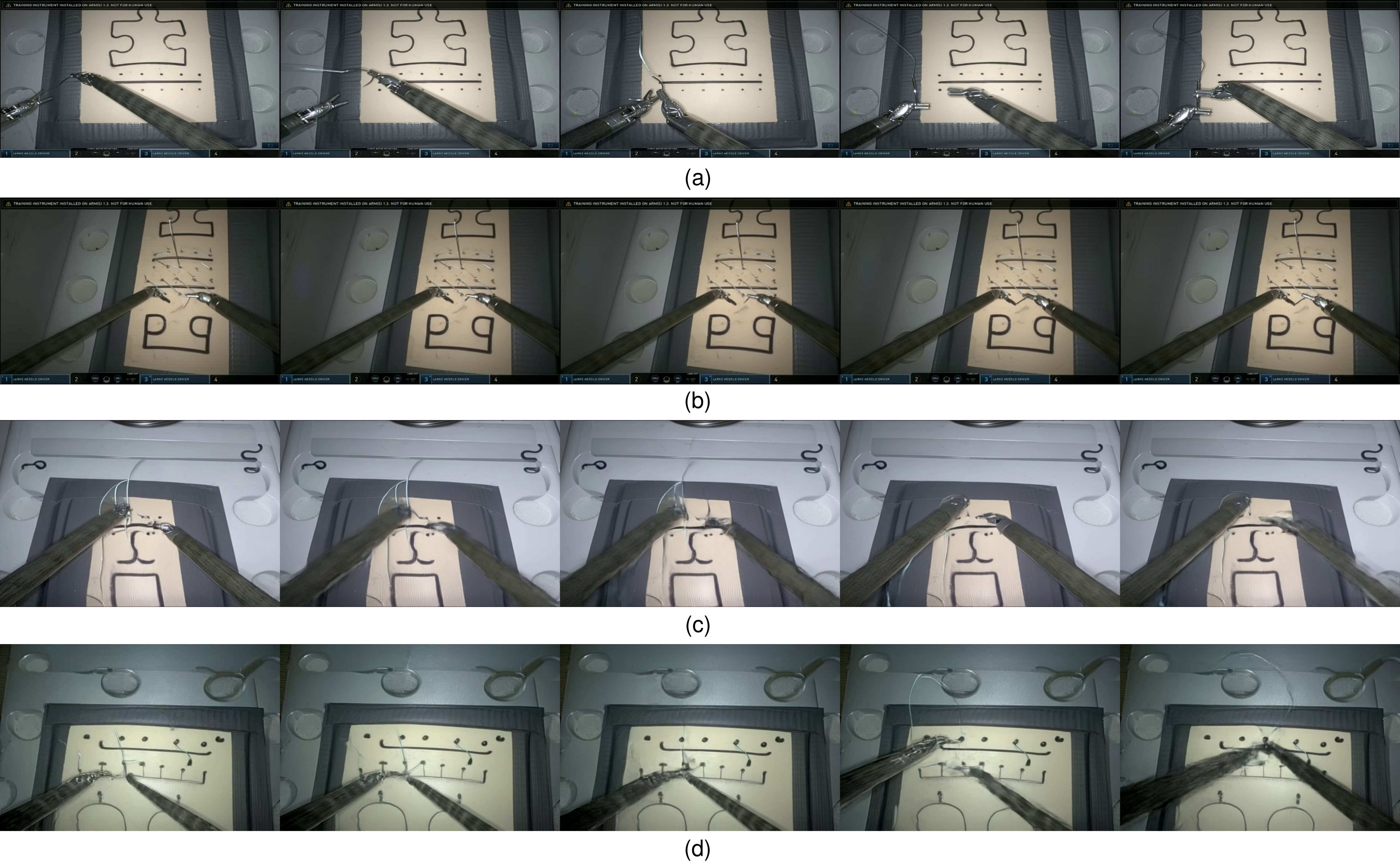} 
  \caption{Sample outputs of different models compared against a real-world needle driving clip from a backhand suturing task. (a) Real-world sample, (b) HunyuanVideo, (c) LTX Video LoRA, (d) LTX Video full training.}
  \label{fig:single-image}
\end{figure*}

\begin{table*}[h!]
    \centering
    \caption{Model architectures and training specifications. Resolution is specified as width ($W$) × height ($H$) × temporal frames ($T$). Training time measured on dual-node A100 (80GB) GPU configuration with batch size 1.}
    \label{tab:model_config}
    \begin{tabular}{l c c c}
        \hline
        \textbf{Model} & \textbf{Training Resolution (W×H×T)} & \textbf{Training Time (Hours)}  \\
        \hline
        LTX-Video (t2v) &  768×512×49 & 15  \\
        [1ex]
        HunyuanVideo (t2v) & 768×512×49 & 71  \\
        [1ex]
        LTX-Video (i2v) & 1024×576×49, 960×444×65, 512×288×121 & 35  \\
        \hline
    \end{tabular}
\end{table*}

\begin{table*}[t!]
    \centering
    \caption{Quantitative performance evaluation of fine-tuned video diffusion models. L2 reconstruction loss calculated between predicted and ground truth pixel values (lower is better). Inference time measured in seconds per video generation (mean of 10 runs) on a single A100 GPU with CFG scaling=3.0 for LTX-Video, and CFG scaling=6.0 for HunyuanVideo for text-to-video models and STG scaling=1.0 for image-to-video.}
    \label{tab:quant_eval}
    \begin{tabular}{l c c}
        \hline
        \textbf{Model} & \textbf{Loss (L2 Reconstruction)} & \textbf{Inference Time (s)}  \\
        \hline
        LTX-Video t2v (LoRA) & 0.32576 & 6.1  \\
        [1ex]
        LTX-Video t2v (Full Finetune) & 0.32928 & 6.1  \\
        [1ex]
        LTX-Video i2v (LoRA) & 0.24501 & 18.7  \\
        [1ex]
        HunyuanVideo (LoRA) & 0.12068 & 182.3  \\
        \hline
    \end{tabular}
\end{table*}

\section{Related Work}

Recent work has begun exploring diffusion models for improved robotic surgical assistance, specifically in imitation learning and surgical scene understanding. In imitation learning, Movement Primitive Diffusion (MPD) combines probabilistic movement primitives with diffusion to enable precise robot actions on deformable tissues \cite{scheikl_movement_2024}. Similarly, the Surgical Robot Transformer (SRT) integrates diffusion policies conditioned on visual feedback with transformers to address kinematic inaccuracies inherent to robotic surgery \cite{kim_surgical_2024}. Complementing these, diffusion policies like the Diffusion Stabilizer Policy (DSP) further improve robustness by automatically denoising imperfect surgical demonstrations, outperforming standard imitation methods \cite{ho_diffusion_2025}.

In parallel, diffusion models have also demonstrated promise for surgical scene synthesis and action planning. Multi-Scale Phase-Conditioned Diffusion (MS-PCD) is a hierarchical diffusion planner capable of goal-directed action sequence prediction from surgical videos \cite{zhao_see_2024}. Generative frameworks such as SimGen and the Anatomy-Aware Diffusion model generate realistic surgical images with accurate anatomical annotations, with the aim of alleviating data scarcity for surgical computer vision tasks \cite{bhat_simgen_2025,venkatesh_data_2025}.  Endora is an endoscopy video simulator that integrates a latent diffusion model with a spatial-temporal transformer and vision foundation model guidance to produce clinical endoscopy videos \cite{li2024endora}. Grasp Anything for Surgery (GAS) presents a world-model-based deep reinforcement learning approach that enables robot grippers to grasp various surgical objects using pixel-level visuomotor policies without requiring pose estimation or feature tracking \cite{lin2024world}. SurgSora leverages RGB-depth feature fusion and optical flow mapping to enable trajectory-guided surgical video synthesis from single frames \cite{chen2024surgsora}. %

Latent video diffusion models have recently shown great promise for generating videos with realistic visuals and motion. For our models, we utilize HunyuanVideo and LTX-Video due to their superior visual quality. LTX-Video presents an optimized model for video generation that streamlines Video-VAE and transformer interactions with a high compression ratio, enhanced positional embeddings, and QK normalization, achieving high visual quality and faster than real-time performance \cite{hacohen2024ltx}.  HunyuanVideo is an open-source 13B-parameter video generative model with flow matching \cite{kong2024hunyuanvideo}. The model excels in motion dynamics and visual quality. %

\section{Methodology}
\subsection{Dataset}
We manually annotated start and end times of sub-stitch actions in 102 training session videos. Each sub-stitch was expert-annotated with a binary technical score (ideal or non-ideal), reflecting the operator’s skill while performing the sub-stitch action.  We created separate clips for each of the actions, and automatically generated captions for them using the expert annotations and the type of task. The final dataset comprises 1,836 video clips from railroad and backhand suturing exercises, annotated with detailed sub-stitch classifications distinguishing between ideal and non-ideal executions of fundamental actions:

\begin{itemize}
\item \textbf{Needle Positioning} (ideal/non-ideal): Grasping and orienting the needle appropriately
\item \textbf{Needle Targeting} (ideal/non-ideal): Approaching the tissue at the correct angle and position
\item \textbf{Needle Driving} (ideal/non-ideal): Passing the needle through tissue with proper wrist rotation
\item \textbf{Needle Withdrawal} (ideal/non-ideal): Extracting the needle along its curved trajectory
\end{itemize}
We use expert-annotated labels to generate example prompts like the following:\vspace{0.1cm}\\
\textit{Example Prompt 1:} \textit{A non-ideal clip of a needle driving action during a backhand task.}\\
\textit{Example Prompt 2:} \textit{An ideal clip of a needle positioning action during a railroad task.}\\
\textit{Example Prompt 3:} \textit{A non-ideal clip of a needle withdrawal action during a backhand task.}

\subsection{Models}
We utilize two open-source video diffusion models: LTX-Video (2B parameters) and HunyuanVideo (13B parameters). Initially, we train text-to-video (t2v) models. Motivated by the positive results, we scale LTX-Video training and train an image-to-video (i2v) model.

\section{Experiments}

\subsection{Training}
We train models using two compute nodes, each with 1x 80GB A100 GPU. We use $rank=256$  LoRAs with $\alpha=256$. %
We train our models for 3 epochs and report training time per model, except for LTX-Video i2v which we train for 30 epochs. 

\subsection{Results}

For inference we use Classifier-Free Guidance (CFG) for t2v models, and Spatiotemporal Skip Guidance (STG) for i2v \cite{hyung2024spatiotemporal}. Our training configurations are summarized in Table~\ref{tab:model_config}. We show sample results from our models in Figure \ref{fig:single-image}. We provide the inference time (mean of 10 runs) and L2 reconstruction loss values for all models in Table~\ref{tab:quant_eval}. Quantitative evaluation shows that the 13B-parameter HunyuanVideo achieves superior reconstruction fidelity (L2 loss: 0.12068) at the cost of significantly increased computational demands (inference latency: 182.3s), presenting deployment constraints for real-time clinical applications. We further find that the i2v LTX-Video model achieves good results while maintaining a low latency.

\section{Conclusion}
We demonstrate the feasibility of latent video diffusion models as surgical world models for robotic suturing that are capable of distinguishing between ideal and non-ideal techniques. Our quantitative evaluation shows promising results for potential deployment of models to the operating room for real-time feedback to surgeons. Future research will focus on (i) integrating kinematic data for multimodal understanding of instrument-tissue interactions, (ii) developing stereo-aware capabilities to better capture depth relationships, and (iii) applying model compression techniques to facilitate real-time clinical deployment. Dataset expansion will enhance generalization across surgical scenarios, while integration with reinforcement learning could enable autonomous/semiautonomous surgical systems. These advances have potential applications in surgical training, assessment, and intraoperative guidance.

{
    \small
    \bibliographystyle{ieeenat_fullname}
    \bibliography{main}
}

\end{document}